# belabBERT: a Dutch RoBERTa-based language model applied to psychiatric classification

Joppe Wouts[1*], Janna de Boer[1,2], Alban Voppel[1], Sanne Brederoo[1,3], Sander van Splunter[4] & Iris Sommer[1]

[1] Department of Biomedical Sciences of Cells & Systems, UMC Groningen, The Netherlands.
[2] Department of Psychiatry, UMC Utrecht Brain Center, UMC Utrecht, The Netherlands
[3] Center for Psychiatry, University Medical Center Groningen, The Netherlands
[4] Informatics Institute, Faculty of Science, University of Amsterdam, The Netherlands
* firstiniatial.secondinitial.lastname@umcg.nl

## Abstract

Natural language processing (NLP) is becoming an important means for automatic recognition of human traits and states, such as intoxication, presence of psychiatric disorders, presence of airway disorders and states of stress. Such applications have the potential to be an important pillar for online help lines, and may gradually be introduced into eHealth modules. However, NLP is language specific and for languages such as Dutch, NLP models are scarce. As a result, recent Dutch NLP models have a low capture of long range semantic dependencies over sentences. To overcome this, here we present belabBERT, a new Dutch language model extending the RoBERTa architecture. belabBERT is trained on a large Dutch corpus (+32 GB) of web crawled texts. We applied belabBERT to the classification of psychiatric illnesses. First, we evaluated the strength of text-based classification using belabBERT, and compared the results to the existing RobBERT model. Then, we compared the performance of belabBERT to audio classification for psychiatric disorders. Finally, a brief exploration was performed, extending the framework to a hybrid text- and audio-based classification. Our results show that belabBERT outperformed the current best text classification network for Dutch, RobBERT. belabBERT also outperformed classification based on audio alone.

## 1. Introduction

Recent advances in natural language processing (NLP) have shown that analysis of spoken language can provide valuable information about a person's mindset. This information can be used for multiple purposes, including the detection of intoxications (for example in traffic), the detection of distress in online services and for use in screening of diagnostics in healthcare. Our group has experience with the use of NLP to support psychiatric diagnosis (de Boer et al. 2020). For such purposes, both acoustic and semantic analyses can provide valuable and complementary information (Corcoran and Cecchi, 2020; Corcoran et al., 2020; de Boer et al., 2020). Semantic space models such as Latent Semantic Analysis (LSA; Landauer, Foltz and Laham, 1998) and word2vec (Mikolov, Sutskever, et al., 2013) are highly promising for this purpose, because they aim to capture coherence or similarity in language, which is often disturbed in schizophrenia spectrum disorders. Indeed, studies to date show that (developing) schizophrenia spectrum disorders and healthy participants can be accurately classified using semantic space models applied to spoken languages (Elvevag et al., 2007; Bedi et al., 2015; Corcoran et al., 2018; Rezaii, Walker and Wolff, 2019; Voppel et al., 2021). Similar studies have been performed for patients with a depressive disorder, who also show marked speech disturbances (Koops et al., 2021). However, most research to date has been performed on the English language, which has the advantage of large available corpora.

For small languages, such as Dutch, pre-trained monolingual NLP models are scarce. As a result, Dutch natural language models have a low capture of long range semantic dependencies over sentences. Here, we present belabBERT, a new Dutch language model extending the RoBERTa architecture. belabBERT is trained on a large Dutch corpus (+32GB) of web crawled texts. In the following, we first discuss the current available Dutch language models, and show how belabBERT performs in comparison to the present models. After we evaluate the strength of text-based classification in a psychiatric sample, we briefly explore the extension of the framework to a hybrid text- and audio-based classification. Using this hybrid framework we show the principle of hybridization with a basic audio-classification network.

## 2. Related work
### 2.1. Semantic space models

A substantial variety of approaches exists in the field of NLP, ranging from tagging parts-of-speech with the aim of finding characterizing patterns in the syntactical representation of text, to building semantic spaces to represent words as mathematical objects. The latter approach has seen a rapid rise in tackling various linguistic problems. A meta-analysis of eighteen studies in which semantic space models are used in psychiatry and neurology supports the notion that analyzing full sentences is more effective than analyzing single words (de Boer et al., 2018). The best performing models were based on word2vec, which uses word embeddings to represent sequences of words in analyzing text (Mikolov, Chen, et al., 2013; Mikolov, Sutskever, et al., 2013). However, word2vec lacks the ability to analyze full sentences or longer-range dependencies. Bidirectional transformer models such as BERT (Devlin et al., 2018) use word embeddings as input similar to word2vec, but have the advantage of handling longer input sequences and take into account the relations within these sequences. This ability, combined with the attention mechanism famously described in the "attention is all you need" paper (Vaswani et al., 2017) enables BERT to find long range dependencies in text, leading to more robust language models. Unsurprisingly, current NLP research is dominated by the use of BERT. All top 10 submissions for the GLUE benchmark make use of BERT models, thus rendering it a suitable candidate model for written text analysis in the current approach. Figure 1 shows a BERT architecture for sentence classification.

**Figure 1.** BERT architecture for sentence classification task

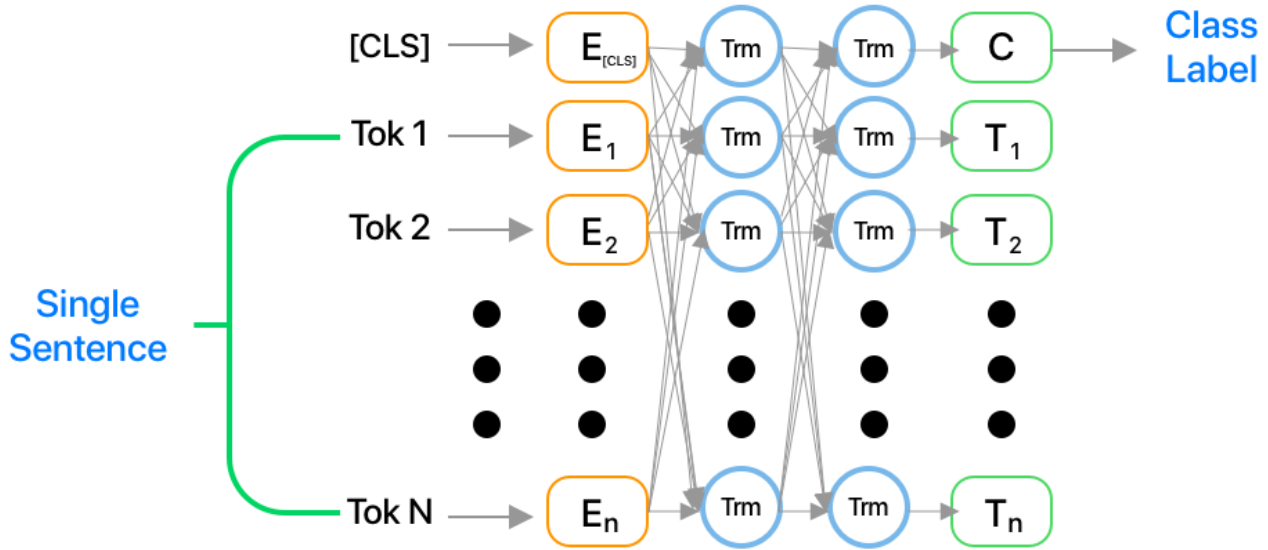

The original BERT model was pre-trained on a large quantity of multilingual data. However, since the open sourcing of the BERT architecture by Google, a multitude of new models have been made available, including monolingual models constructed for tasks in specific languages (Kuratov and Arkhipov, 2019; Martin et al., 2019; Virtanen et al., 2019; Antoun, Baly and Hajj, 2020).

A comparison of monolingual BERT model performance and multilingual BERT model performance on various tasks showed that monolingual BERT models outperform multilingual models on every task (Nozza, Bianchi and Hovy, 2020), see Table 1.

The top performing model for Dutch is RobBERT (Delobelle, Winters and Berendt, 2020) which is a BERT model using a different set of hyperparameters dubbed RoBERTa (Liu et al., 2019). Another model, BERTje (de Vries et al., 2019), is more traditional in the sense that the pre-training hyperparameters follow the parameters as described in the original BERT publication.

**Table 1.** Different monolingual BERT model average performance versus multilingual BERT model performance on various tasks

| Task | Metric | Avg. Monolingual BERT | Avg. Multilingual BERT | Diff |
|---|---|---|---|---|
| Sentiment Analysis | Accuracy | **90.17** % | 83.80 % | 6.37 % |
| Text Classification | Accuracy | **88.96** % | 85.22 % | 3.75 |

## 2.2. Audio classification

Spontaneous speech contains a wealth of information that reflects personal characteristics of the speaker such as mood, motivation, intelligence, motor control (i.e. articulation) and arousal. Related work in audio analysis for diagnostic purposes found that impressive results can be achieved using speech parameters alone (Cohen, Mitchell and Elvevåg, 2014; Martínez-Sánchez et al., 2015; Parola et al., 2019; Tahir et al., 2019). The majority of research in this field focuses on more traditional machine learning techniques such as logistic regression or support vector machine. However, these are less resistant to noise in the data and thus require feature engineering before processing the parameters. A notable weakness of feature engineering is that information is lost, as it is difficult for traditional machine learning techniques to cope with noise that irrelevant features provide.

## 2.3. Current project

Here, we present belabBERT, a new Dutch language model extending the RoBERTa architecture. We first discuss the current Dutch language models, and show how belabBERT performs in comparison to the present models. We apply the belabBERT model to psychiatric classification, after which we briefly explore hybrid text- and audio-based classification.

## 3. Methods
## 3.1. belabBERT

All experiments were run on a high-performance computing cluster. The language model belabBERT was trained on 16 NVidia Titan RTX GPUs (24 GB RAM each) for a total of 60 hours. All other tasks were run on a single node containing 4 GPUs of the same specifications. We use the original RoBERTa training parameters. In order to fine-tune belabBERT and RobBERT for the classification of text input we implemented the classifier head as described in the BERT paper. In order to find the optimal hyperparameter set we performed several runs with different sets of configurations.

### 3.1.1. Pre-training

For the pre-training of belabBERT we used the OSCAR corpus which consists of a set of monolingual corpora extracted from Common Crawl snapshots. A non-shuffled version was made available for the Dutch corpus, which consists of 41 GB of raw text. This is in contrast with the corpus used for RobBERT, which uses the shuffled and pre-cleaned version. By using a non-shuffled version the sentence order of the corpus is preserved. This property was chosen with the aim of enabling belabBERT to learn long range syntactic dependencies. RobBERT uses the RoBERTa base tokenizer which is a tokenizer trained on an English corpus; we assume this affects the performance of RobBERT negatively on downstream tasks. We therefore trained a byte pair encoding tokenizer on the same OSCAR corpus to create the word embeddings which belabBERT uses as input, alleviating potential problems in RobBERT for both regarding tokenizer as well as long-term dependencies. Additionally, we performed a sequence of common preprocessing steps in order to better match the our psychiatry interview transcript data. These preprocessing steps included fuzzy deduplication (i.e. removing lines with a >90% overlap with other lines), removing non-textual data such as "https://", and excluding lines longer than 2000 words. This resulted in a total amount of 32 GB clean text of which 10% was held-out as validation set to accurately measure overfitting.

### 3.1.2. Coding

The language model belabBERT was created using the Hugging Face's transformer library, a Python library which provides a lot of boilerplate code for building BERT models. belabBERT uses a RoBERTa architecture and unless otherwise specified all parameters for the training of this model were kept default.

The model and used code is publicly available under an MIT open-source license on GitHub[1]. All other models used (text classifier, audio classifier and hybrid classifier) were developed in Python using the PyTorch Lightning framework. Hyperparameter optimization was performed using the Weights & Biases Sweeps system. This process involves generating a large set of configuration parameters based on pre-defined default parameter values and training the model accordingly. We picked the model with the lowest cross-entropy loss on the held-out validation set, on the assumption that this model is most generalizable.

### 3.1.3. Training

A Dutch BPE tokenizer was used for belabBERT to create its word embeddings, which makes it an efficient tokenizer for our dataset when compared to the Multi lingual tokenizer used for RoBERTa. As a consequence, belabBERT produces fewer tokens for a Dutch text than RobBERT, which explains the skewed sizes of training samples. Our default hyperparameters follow the GLUE fine-tuning parameters used in the original RoBERTa paper. Subsection 3.3 shows the training configuration which was used for the hybrid model. This involves two neural networks which were trained separately, in which the first described model takes audio features as input, and the second is the fusion layer which bases its output classification on 6 tensorized input values. In order to find the optimal set of hyperparameters we trained each model 15 times. We show the parameter set for the described model that reached the lowest cross-entropy validation loss.

## 3.2. Acoustic model

Using a neural network enables us to use all audio extracted speech parameters as input and automatically learn which features are relevant for each classification.

The held-out test set of our audio classifier consisted of all samples of which a transcript did exist, ensuring there is no overlap between the training data of the audio classifier and the text classifier. The audio classification network uses categorical cross-entropy loss and Adam optimization (Kingma and Ba, 2014) with $\beta_1 = 0.9, \beta_2 = 0.95$ and $\epsilon = 10^{-8}$. Due to the inherent noisy nature of an audio signal and its extracted features we used a default dropout rate of 0.1. The learning rate boundaries were found by performing an initial training run, during which the learning rate linearly increases for each epoch (Smith, 2017). We picked the median learning rate of these bounds as our default learning rate.

## 3.3. Hybrid model

We developed a hybrid model making use of both modalities (text and audio) and compared its performance to the single models. We assume this model improves the accuracy of the classification since audio characteristics are not embedded in text data; e.g. variations in pitch can be highly indicative for depression, but are not present in text data. Similarly, coherence of grammar and semantic dependencies are indicative of the mental state of a person, but cannot be not found in the audio signal. There are multiple ways and techniques to combine models. As the current approach aims to present an initial proof of concept for hybridization, we chose a simple "late fusion" architecture with a fully-connected layer to map the output of both models into 3 outputs. After training both models separately, weights were frozen and output layers of the separate models were used to generate inputs for the hybrid model.

---

[1] https://github.com/iPieter/RobBERT/blob/master/notebooks/finetune_dbrd.ipynb

## 4. Evaluation of the models

We first evaluated belabBERT on text sentiment analysis accuracy . Then we tested its application value by evaluating the previously described models on a sample of psychiatric patients.

### 4.1. Sentiment analysis

We replicated the high-level sentiment analysis task used to evaluate BERT-NL (Brandsen et al., 2019) and BERTje (de Vries et al., 2019) to be able to compare our methods. This task uses a dataset called Dutch Book Reviews dataset (DBRD), in which book reviews from hebban.nl are labeled as positive or negative (Van der Burgh and Verberne, 2019). Table 2 provides a short overview of the current Dutch BERT models, in comparison to the belabBERT model presented here.

**Table 2.** The 3 top performing monolingual Dutch BERT models based on their sentiment analysis accuracy

| Model name | Pre-train corpus | Tokenizer type | Acc Sentiment analysis |
|---|---|---|---|
| belabBERT | Common Crawl Dutch (non-shuffled) | BytePairEncoding | 95.92* % |
| RobBERT | Common Crawl Dutch (shuffled) | BytePairEncoding | 94.42 % |
| BERTje | Mixed (Books, Wikipedia, etc) | Wordpiece | 93.00 % |

* to be verified

### 4.2. Psychiatric dataset

A total of 339 participants, of which 170 patients with a schizophrenia spectrum disorder, 22 diagnosed with depression, and 147 healthy controls were included in this study. A visualization of the model architecture for psychiatric classification can be found in Figure 2.

**Figure 2.** Model architecture for psychiatric classification, green text marks regular tokens and pink text marks the special tokens indicating a begin of sentence token

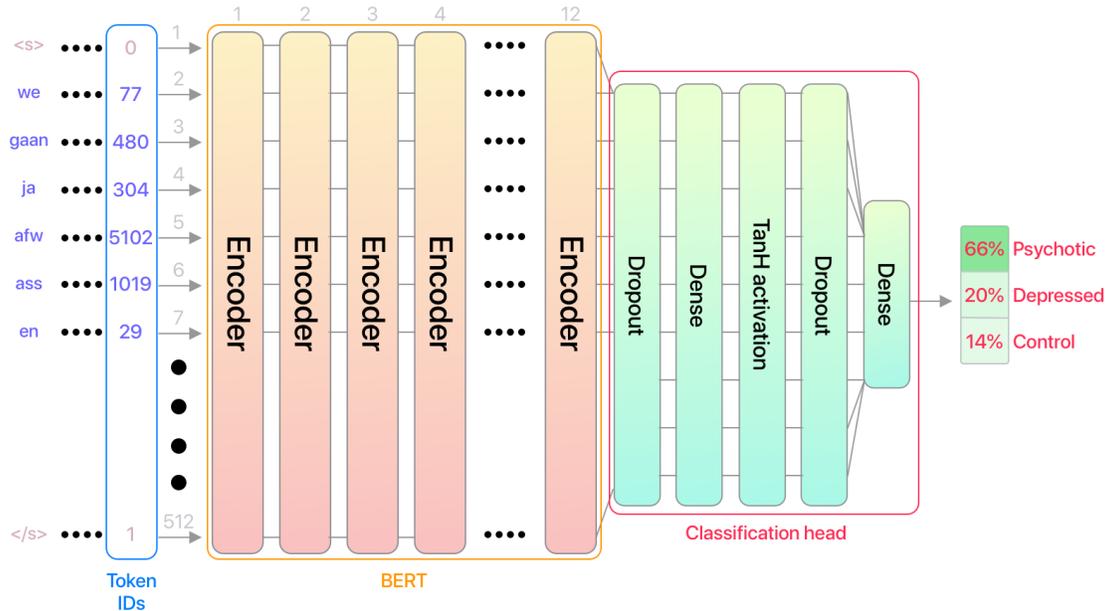

Speech samples were collected from all participants, using a semi-structured interview to elicit spontaneous speech. AKG-C544l head-worn cardioid microphones were used to record the speech. Speech was digitally recorded onto a Tascam DR40 solid state recording device at a sampling rating of 44,100 kHz with 16-bit quantization. The interviewers were trained to avoid health related topics in order to make produced language by the participants more generalizable irrespective of diagnosis or absence thereof.

      The OpenSMILE audio processing framework was used, employing the eGeMAPS parameter set to extract speech parameters for each audio file (Eyben et al., 2016). Of the 339 interviews, 141 were manually transcribed, of which 76 were from psychotic, 6 from depressive, and 59 from healthy participants. Transcripts were made according to the CHAT transcription format (MacWhinney, 2017) by trained transcribers. Transcripts were transformed from the CHAT format to flat text. Since we are dealing with privacy-sensitive information we took measures to mitigate any risk of leaking sensitive info. For audio, we only performed analyses on parameters that were derived from the raw audio, not including any content. For the transcripts, we swapped all transcripts with their tokenized versions and only performed calculations on these. In order to create more examples, full tokenized transcripts were chunked into a length of 220 tokens per chunk and 505 tokens per chunk, resulting in two transcript datasets per tokenizer. Table 3 shows the amount of samples after chunking. The acquired datasets were split into 80% training set, 10 % validation and 10 % test set keeping the ratios among participants of the original dataset.

Table 3. Total amount of samples after chunking with different chunk lengths and different tokenizers

| Dataset ID | Chunk size | Psychotic | Control | Depressive | Total |
|---|---|---|---|---|---|
| belabBERT-220 | 220 | 625 | 589 | 52 | 1266 |
| belabBERT-505 | 505 | 294 | 274 | 24 | 592 |
| RobBERT-220 | 220 | 1096 | 1043 | 92 | 2231 |
| RobBERT-505 | 505 | 499 | 127 | 41 | 1012 |
| **Full** | — | 76 | 59 | 6 | 141 |

## 4.3. Results
### 4.3.1. belabBERT

In order to evaluate the performance of belabBERT we evaluated it against the performance of the current Dutch state-of-the-art model RobBERT. The results of these experiments help to better contextualize the achieved results of belabBERT. To measure the effect of chunk sizes we ran two separate analyses for each base model (belabBERT and RobBERT), with a varying chunk size of 220 and 505 (Tables 4a and 4b), tested for each model.

Table 4a. Overview of samples per category for training belabBERT and RobBERT

| Dataset ID | Category | Train set | Validation set | Test set |
|---|---|---|---|---|
| belabBERT-220 | Psychotic | 500 | 62 | 63 |
|  | Control | 471 | 59 | 59 |
|  | Depressed | 41 | 5 | 6 |
| belabBERT-505 | Psychotic | 235 | 29 | 30 |
|  | Control | 219 | 27 | 28 |
|  | Depressed | 19 | 2 | 3 |
| RobBERT-220 | Psychotic | 876 | 110 | 110 |
|  | Control | 834 | 104 | 105 |
|  | Depressed | 73 | 9 | 10 |
| RobBERT-505 | Psychotic | 398 | 50 | 51 |
|  | Control | 100 | 13 | 14 |
|  | Depressed | 31 | 5 | 5 |

**Table 4b.** Parameters for best performing model for belabBERT and RobBERT

| Dataset ID | Batch size | Epochs | Peak learning rate | Warmup steps |
|---|---|---|---|---|
| belabBERT-220 | 9 | 5 | $8.42e^{-5}$ | 190 |
| belabBERT-505 | 10 | 3 | $6.22e^{-5}$ | 373 |
| RobBERT-220 | 13 | 3 | $6.58e^{-5}$ | 401 |
| RobBERT-505 | 10 | 3 | $1.19e^{-4}$ | 401 |

Table 5 and Figure 3 show that both experiments with belabBERT as its base model manage to outperform the current Dutch state-of-the-art RobBERT, with the top performing model using a chunk size of 220 achieving a classification accuracy of 75.68% on the test set and 71.18% on validation set. The top performing model with RobBERT as base also uses a chunk size of 220 and reaches a 69.06% classification accuracy on the test set and 69.64% on the validation set.

**Table 5.** Classification accuracy for the best performing belabBERT and RobBERT based models on the held-out validation and test set

| Experiment | Validation accuracy | Test accuracy |
|---|---|---|
| belabBERT-220 | **71.18%** | **75.68%** |
| belabBERT-505 | 70.25% | 73.91% |
| RobBERT-220 | 69.64% | 69.06% |
| RobBERT-505 | 68.93% | 65.69% |

**Figure 3.** Results for the belabBERT based model with a chunk size of 220: Predicted classes vs actual classes

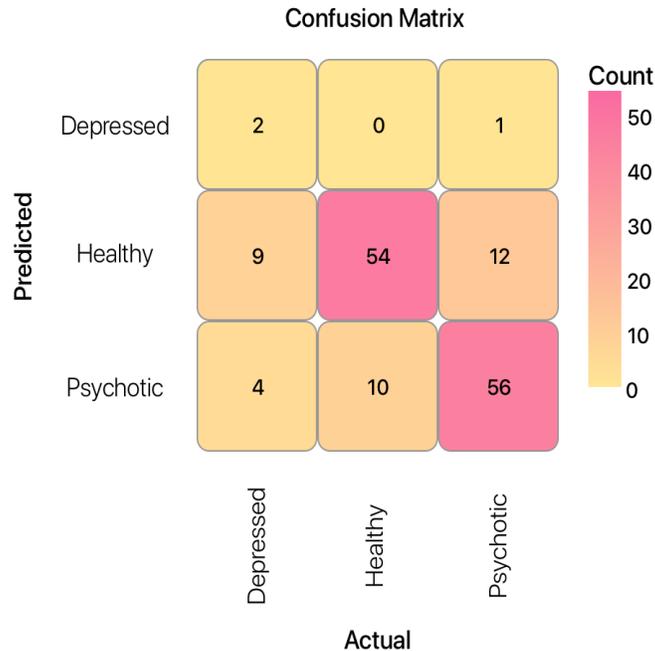

belabBERT indeed benefits from its ability to capture long range semantic dependencies. Both on the 505 chunk size, as well as the 220 chunk size experiments belabBERT manages to outperform the current state-of-the-art language model RobBERT. belabBERT 220 has a limited recall for the depression label but its precision is higher than expected (Table 6).

**Table 6.** Classification metrics for the belabBERT based model with a chunk size of 220

| Metric | Depressed | Healthy | Psychotic |
|---|---|---|---|
| Recall | 13.33% | **84.38%** | 81.16% |
| Precision | 66.67% | 72.00% | **80.00%** |
| F1-score | 22.21% | 77.70% | **80.58%** |

### 4.3.2. Audio classification

We used a simple neural network architecture consisting of three layers of which the specifics can be seen in Figure 4.

**Figure 4.** Model architecture for audio classification based on extracted speech features

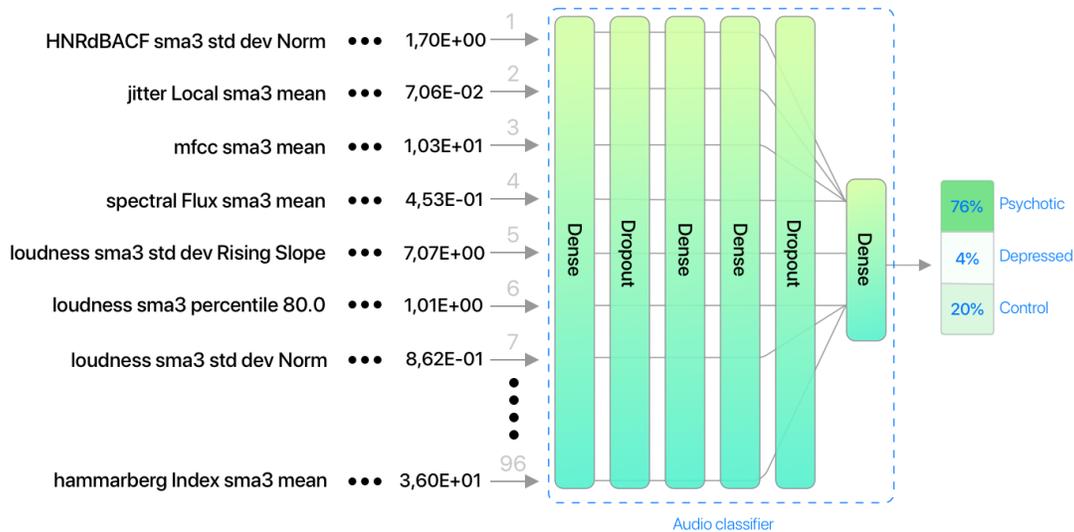

In order to maximize the size of available training samples for the fusion we trained the audio classifier on samples of which no transcript were available (see Table 7).

**Table 7**. Overview of samples per category for training Audio classification network

| Set | Psychotic | Control | Depressed | % Of total |
|---|---|---|---|---|
| Train | 97 | 74 | 7 | 53 |
| Validation | 10 | 8 | 2 | 6 |
| Test | 76 | 59 | 6 | 41 |
| Total | 183 | 141 | 15 | 100% |

The audio classification network reached a classification accuracy of 65.96 % on the test set and 80.05% accuracy on the validation set. Due to the small size of this set we should not consider this result as significant. In addition, the network was not able to distinguish samples with the depressed label from the other labels based on its inputs (Figure 5 and Table 8).

**Figure 5.** Audio classifier results: Predicted classes vs actual classes

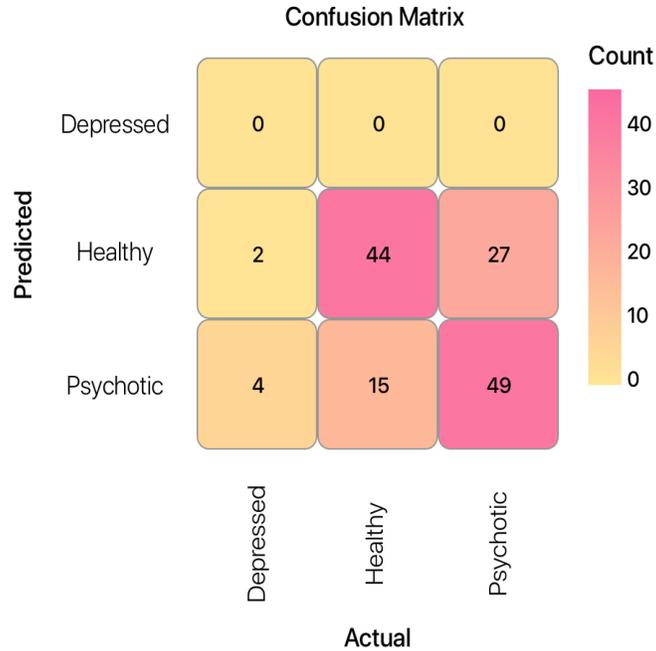

**Table 8.** Classification metrics for audio classification

| Metric | Depressed | Healthy | Psychotic |
| --- | --- | --- | --- |
| Recall | 0% | **75.6%** | 64.47% |
| Precision | 0% | 60.27% | **72.05%** |
| F1-score | 0% | 67.07% | **68.05%** |

### 4.3.3. Hybrid classification

We trained the hybrid classification model (see Figure 6) on the dataset of our best performing text classification network (belabBERT-220). It is of note that due to the chunking of this dataset we have multiple samples stemming from single patients. As discussed in section 4.2, this explains the difference in total amount of samples between the audio classification and hybrid classification.

**Figure 6.** Hybrid model architecture for classification task

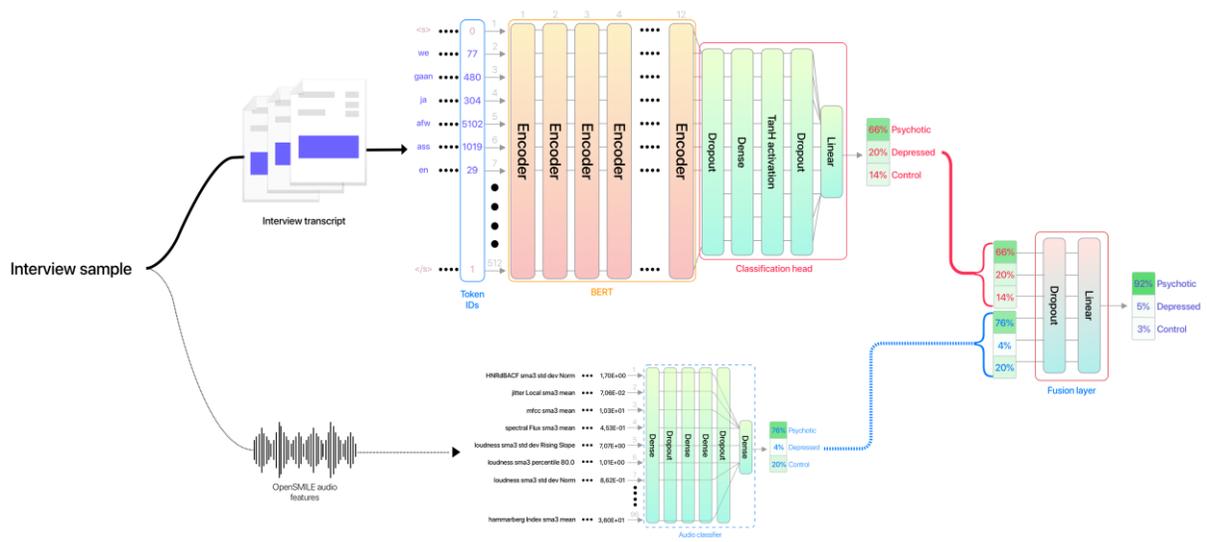

The classification accuracies for the hybrid classification network reached an accuracy of 77.70% on the test set and a 70.47% accuracy on the validation set (see also Figure 7 and Table 9).

**Figure 7.** Hybrid classifier results: Predicted classes vs actual classes

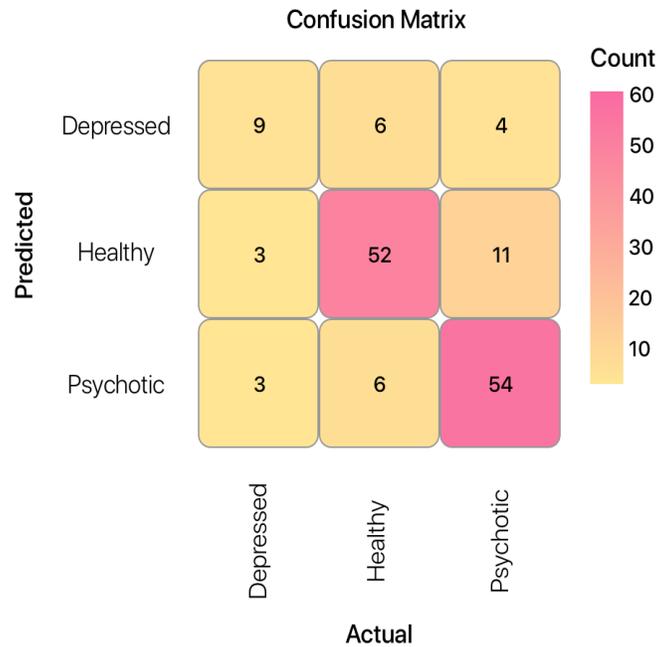

**Table 9.** Classification metrics for hybrid classification

| Metric | Depressed | Healthy | Psychotic |
|---|---|---|---|
| Recall | 60.00% | **81.25%** | 78.26% |
| Precision | 47.37% | 78.79% | **85.71%** |
| F1-score | 52.94% | 80.01% | **81.82%** |

We can conclude that the hybrid classification model performs relatively well on the healthy and the psychotic category, but not on the depressed label. The extension towards the hybrid model where we base our classification on both text and audio does, however, result in an improved classification accuracy over the text or audio alone classifiers.

## 5. Discussion

We constructed a new lexicon based on a large dataset of web crawled texts and compared the new model to the existing one. Our belabBERT model reached a 6.62% higher accuracy in classifying psychiatric diagnoses compared to the best performing RobBERT model (Table 5). Furthermore, we observe that a smaller chunk size of 220 tokens lead to a significant accuracy gain over the 505 chunk size. The small difference between the validation and test set accuracies shown in Table 5 are a positive indicator that the classification accuracy is significant and representative for the capability of the model to categorize the text samples. Given that we found that belabBERT outperforms RobBERT in classification accuracy, we conclude that using a specialized Dutch tokenizer and pre-training corpus containing conversational data improves downstream classification tasks. In addition, we showed that using a smaller chunk size has a positive effect on the classification accuracy.

Our brief exploration into the hybridization of belabBERT with a basic audio-classification network increased accuracy of 75.68% to a 77.0% accuracy. From our observations of the classification metrics shown in Table 9 we showed that the addition of an audio classification network to the strong stand-alone text classification model leads to an overall better precision for all labels, as well as to higher classification accuracy. However, the small group of 'depressed' speech samples in our dataset prevents definitive conclusions about relevance of our findings in this category.

### 5.1. Future work

Compared to BERT models of the same size as belabBERT, it seems that belabBERT is in fact still under-trained. The version used in the current approach has only seen 60% of the training data. Training belabBERT even more could possibly increase its performance on all tasks. In our text classification we already applied a chunking technique in order to generate more examples from a single interview sample. However, we observed that prediction accuracy increased when we decreased the chunk size. This suggests there may be merit in exploring how the use of even smaller chunk sizes affects prediction accuracy. When smaller chunk sizes can be used, the amount of training examples is increased, making the model more robust.

While the explored hybrid model we present here used pre-extracted audio parameters as input for a neural network, it would be interesting to apply new audio analysis techniques, for example by using raw audio as input for a neural network. The approach would be similar to speech recognition architectures; a major advantage would be that these architectures can find patterns over time, which enables the discovery of new relations between input features. The hybrid model could also use other data sources to generate a classification, such as video, which would possibly increase classification accuracy even further.

The interpretation and rationalization of the predictions of neural networks is key for providing clinical relevancy, not only in the practical domain of psychiatry but also for the theoretic understanding of the disorder and symptoms. Transformer models like BERT are easily visualized, and an extensive interpretation toolkit could provide researchers with better tools to discover new patterns in language that are highly indicative for a certain classification prediction, in turn increasing our understanding of the disorders.

## 5.2. Conclusion

We presented a strong classification model for the Dutch language. This new model, "belabBERT", is trained on capturing long range semantic dependencies over sentences in a Dutch corpus. We apply this model to text classification and show that this language model outperforms the current state-of-the-art RobBERT model. We, furthermore, showed that we could increase the size of our dataset by splitting the samples up into chunks of a fixed length without losing classification accuracy. In addition, we explored the possibilities for a hybrid network which uses both text and audio data as input for the classification of individuals as psychotic or depressed patients, or non-clinical. Our results indicate this approach is indeed able to improve the accuracy and precision of a standalone text classification network. Based on these observations we can confirm our main hypothesis that a well-designed text-based approach poses a strong competition against the state-of-the-art audio based approaches for the classification of psychiatric illness.

## 6. References

Antoun, W., Baly, F. and Hajj, H. (2020) 'Arabert: Transformer-based model for arabic language understanding', *arXiv preprint arXiv:2003.00104*.
Bedi, G. *et al.* (2015) 'Automated analysis of free speech predicts psychosis onset in high-risk youths.', *NPJ schizophrenia*, 1(May), p. 15030.
de Boer, J. N. *et al.* (2018) 'Clinical use of semantic space models in psychiatry and neurology: A systematic review and meta-analysis', *Neuroscience & Biobehavioral Reviews*.
de Boer, J. N. *et al.* (2020) 'Anomalies in language as a biomarker for schizophrenia', *Current Opinion in Psychiatry*, 33(3), pp. 212–218.
Brandsen, A. *et al.* (2019) 'BERT-NL a set of language models pre-trained on the Dutch SoNaR corpus'.
Van der Burgh, B. and Verberne, S. (2019) 'The merits of Universal Language Model Fine-tuning for Small Datasets--a case with Dutch book reviews', *arXiv preprint arXiv:1910.00896*.
Cohen, A. S., Mitchell, K. R. and Elvevåg, B. (2014) 'What do we really know about blunted vocal affect and alogia? A meta-analysis of objective assessments', *Schizophrenia Research*, 159(2–3), pp. 533–538. doi: 10.1016/j.schres.2014.09.013.
Corcoran, C. M. *et al.* (2018) 'Prediction of psychosis across protocols and risk cohorts using automated language analysis', *World Psychiatry*, 17(1), pp. 67–75. doi: 10.1002/wps.20491.
Corcoran, C. M. *et al.* (2020) 'Language as a biomarker for psychosis: A natural language processing approach', *Schizophrenia Research*.
Corcoran, C. M. and Cecchi, G. (2020) 'Using language processing and speech analysis for the identification of psychosis and other disorders', *Biological Psychiatry: Cognitive Neuroscience and Neuroimaging*.
Delobelle, P., Winters, T. and Berendt, B. (2020) 'Robbert: a dutch roberta-based language model', *arXiv preprint arXiv:2001.06286*.

Devlin, J. *et al.* (2018) 'Bert: Pre-training of deep bidirectional transformers for language understanding', *arXiv preprint arXiv:1810.04805*.

Elvevag, B. *et al.* (2007) 'Quantifying incoherence in speech: An automated methodology and novel application to schizophrenia', *Schizophrenia Research*. Edited by A. Allen Andreasen, Andreasen, Bamberg, Barch, Bleuler, Bokat, Bowie, Breier, Clark, Cuesta, DeLisi, Docherty, Foltz, Harrow, Harrow, Hoffman, Holm, Jastak, Jurafsky, Kraepelin, Landauer, Landauer, Landauer, Lorch, Manschreck, McKenna, Niznikiewicz, Overall, 93(1–3), pp. 304–316. doi: 10.1016/j.schres.2007.03.001.

Eyben, F. *et al.* (2016) 'The Geneva Minimalistic Acoustic Parameter Set (GeMAPS) for Voice Research and Affective Computing', *IEEE Transactions on Affective Computing*, 7(2), pp. 190–202.

Kingma, D. P. and Ba, J. (2014) 'Adam: A method for stochastic optimization', *arXiv preprint arXiv:1412.6980*.

Koops, S. *et al.* (2021) 'Speech as a biomarker for depression', *CNS & Neurological Disorders - Drug Targets*, In press.

Kuratov, Y. and Arkhipov, M. (2019) 'Adaptation of deep bidirectional multilingual transformers for russian language', *arXiv preprint arXiv:1905.07213*.

Landauer, T., Foltz, P. and Laham, D. (1998) 'An introduction to latent semantic analysis', *Discourse Processes*, 25(October), pp. 259–284. doi: 10.1080/01638539809545028.

Liu, Y. *et al.* (2019) 'Roberta: A robustly optimized bert pretraining approach', *arXiv preprint arXiv:1907.11692*.

MacWhinney, B. (2017) 'Tools for analyzing talk part 1: The chat transcription format'. Carnegie.

Martin, L. *et al.* (2019) 'Camembert: a tasty french language model', *arXiv preprint arXiv:1911.03894*.

Martínez-Sánchez, F. *et al.* (2015) 'Can the Acoustic Analysis of Expressive Prosody Discriminate Schizophrenia?', *The Spanish journal of psychology*, 18(2015), p. E86.

Mikolov, T., Sutskever, I., *et al.* (2013) 'Distributed representations of words and phrases and their compositionality', in *Advances in neural information processing systems*, pp. 3111–3119.

Mikolov, T., Chen, K., *et al.* (2013) 'Efficient Estimation of Word Representations in Vector Space', *Arxiv*, pp. 1–12.

Nozza, D., Bianchi, F. and Hovy, D. (2020) 'What the [mask]? making sense of language-specific BERT models', *arXiv preprint arXiv:2003.02912*.

Parola, A. *et al.* (2019) 'Voice patterns in schizophrenia: A systematic review and Bayesian meta-analysis', *Schizophrenia Research*.

Rezaii, N., Walker, E. and Wolff, P. (2019) 'A machine learning approach to predicting psychosis using semantic density and latent content analysis', *npj Schizophrenia*, 5(1).

Smith, L. N. (2017) 'Cyclical learning rates for training neural networks', in *2017 IEEE winter conference on applications of computer vision (WACV)*. IEEE, pp. 464–472.

Tahir, Y. *et al.* (2019) 'Non-verbal speech cues as objective measures for negative symptoms in patients with schizophrenia', *PLoS ONE*, 14(4), pp. 1–17.

Vaswani, A. *et al.* (2017) 'Attention is all you need', *arXiv preprint arXiv:1706.03762*.

Virtanen, A. *et al.* (2019) 'Multilingual is not enough: BERT for Finnish', *arXiv preprint arXiv:1912.07076*.

de Vries, W. *et al.* (2019) 'Bertje: A dutch bert model', *arXiv preprint arXiv:1912.09582*.